%% file: sample-sigconf.tex
\documentclass[conference]{IEEEtran}
\IEEEoverridecommandlockouts
\usepackage{cite}
\usepackage{amsmath,amssymb,amsfonts}
\usepackage{graphicx}
\usepackage{textcomp}
\usepackage{xcolor}

\usepackage{algorithm,algpseudocode}
\usepackage{mathtools}
\usepackage{subcaption}
\usepackage{leftidx}
\usepackage{booktabs} 
\usepackage{todonotes}

\def\BibTeX{{\rm B\kern-.05em{\sc i\kern-.025em b}\kern-.08em
    T\kern-.1667em\lower.7ex\hbox{E}\kern-.125emX}}
\begin{document}


\title{Towards Optimized Tensor Code Generation for Deep Learning on Sunway Many-Core Processor}


\author{
\IEEEauthorblockN{Mingzhen Li$^{\dag1,2}$, Changxi Liu$^{\dag3}$, Jianjin Liao$^{1}$, Xuegui Zheng$^{1}$, Hailong Yang$^*$$^{1,2}$ \\Rujun Sun$^{4}$, Jun Xu$^{5}$, Lin Gan$^{6}$, Guangwen Yang$^{6}$, Zhongzhi Luan$^{1}$ and Depei Qian$^{1}$}
\IEEEauthorblockA{
School of Computer Science and Engineering, Beihang University$^{1}$, Beijing, China\\
State Key Laboratory of Software Development Environment$^{2}$, Beijing, China\\
National University of Singapore$^{3}$, Singapore\\
State Key Laboratory of Mathematical Engineering and Advanced Computing$^{4}$, Wuxi, China\\
Science and Technology on Special System Simulation Laboratory Beijing Simulation Center$^{5}$, Beijing, China\\
Department of Computer Science and Technology, Tsinghua University$^{6}$, Beijing, China	
}}

\maketitle
\IEEEpeerreviewmaketitle

\input{abstract}


\def\thefootnote{$\dag$}\footnotetext{Contributed equally.}\def\thefootnote{\arabic{footnote}}
\def\thefootnote{$*$}\footnotetext{Corresponding author.}\def\thefootnote{\arabic{footnote}}

\input{introduction}
\input{background}
\input{overview}
\input{aot}
\input{tensor}

\input{evaluation}
\input{relatedwork}
\input{conclusion}

\bibliographystyle{IEEEtran}
\bibliography{sample-sigconf}

\input{appendix}

\end{document}

%% file: abstract.tex
\begin{abstract}
The flourish of deep learning frameworks and hardware platforms has been demanding an efficient compiler that can shield the diversity in both software and hardware in order to provide application portability. Among the existing deep learning compilers, TVM is well known for its efficiency in code generation and optimization across diverse hardware devices. In the meanwhile, the Sunway many-core processor renders itself as a competitive candidate for its attractive computational power in both scientific computing and deep learning workloads. This paper combines the trends in these two directions. Specifically, we propose \textit{swTVM} that extends the original TVM to support ahead-of-time compilation for architecture requiring cross-compilation such as Sunway. In addition, we leverage the architecture features during the compilation such as core group for massive parallelism, DMA for high bandwidth memory transfer and local device memory for data locality, in order to generate efficient codes for deep learning workloads on Sunway. The experiment results show that the codes generated by \textit{swTVM} achieves 1.79$\times$ on average compared to the state-of-the-art deep learning framework on Sunway, across six representative benchmarks. 
This work is the first attempt from the compiler perspective to bridge the gap of deep learning and Sunway processor particularly with productivity and efficiency in mind. We believe this work will encourage more people to embrace the power of deep learning and Sunway many-core processor.
\end{abstract}

%% file: introduction.tex
\section{Introduction}
\label{sec:introduction}

Currently, deep learning has achieved outstanding performance in many fields, including self-driving car~\cite{bojarski2016end}, face detection~\cite{zhang2016joint} and machine translation~\cite{cho2014learning}. The deep learning frameworks such as TensorFlow~\cite{abadi2016tensorflow}, PyTorch~\cite{ketkar2017introduction}, MxNet~\cite{chen2015mxnet}, and Caffe~\cite{jia2014caffe}, provide an efficient platform to support the research and development on intelligent applications. In the meanwhile, emerging deep learning algorithms exhibit increasing demands for massive computation power. To satisfy the computation demand, various accelerating hardwares such as GPU, FPGA~\cite{wang2016dlau} and ASIC~\cite{jouppi2017datacenter} have been applied in the deep learning field. Current deep learning frameworks almost rely on the high performance libraries such as cuDNN~\cite{chetlur2014cudnn} and MKL~\cite{wang2014intel}, which are provided by the hardware vendors to accelerate the deep learning workloads. With new deep learning algorithms and hardwares arising rapidly, the engineering cost for porting the algorithms to the hardwares has increased dramatically. It is necessary to find a way to deploy these emerging deep learning algorithms on the underlying hardwares automatically and efficiently.

To address the above problem, the end-to-end compilers~\cite{rotem2018glow,cyphers2018intel,vasilache2018tensor,chen2018tvm,tiramisu} for deep learning workloads have been proposed. For example, TVM~\cite{chen2018tvm}, XLA~\cite{abadi2016tensorflow}, Tiramisu~\cite{tiramisu} and Tensor Comprehension~\cite{vasilache2018tensor} are the state-of-the-art deep learning compilers. Taking TVM for example, it digests deep learning models implemented using different frameworks as input, and generates efficient model codes targeting various hardware devices as output. Fundamentally, TVM adopts the design of two level optimization to automatically generate codes for deep learning models. On graph level, it applies multiple optimizations to the computation graph derived from the deep learning model, such as operator fusion and data layout transformation. On operator level, it converts the computations into the tensor operations targeting the various hardwares and hides the memory latency by optimizing the instruction pipeline. Moreover, TVM can optimize the code generation automatically according to the shape and data layout of the input to each layer for better performance.

Meanwhile, for its compelling computation power, Sunway many-core processor serves as the basic building block of Sunway TaihuLight supercomputer, which is the first supercomputer to achieve over 100 petaFlops in the world. The Sunway SW26010 processor consists of four core groups (CG). Each CG, including a Management Processing Element (MPE) and 64 Computing Processing Elements (CPEs), can achieve 765 GFlops peak performance in double-precision. The memory attached to each CG is 8GB with the bandwidth of 34.1GB/s. The MPE is a complete 64-bit RISC core, typically used for task control and management, whereas the CPE is also a 64-bit RISC core but with limited functionalities, typically used for computation. In addition, each CPE has a 64KB local device memory (LDM), that is managed explicitly by software. The executables on Sunway are generated through cross-compilation with MPE and CPE as different compilation targets. Due to the limitation of Sunway customized operating system, the dynamic linked libraries are not supported.

To embrace the advantage of automatic compilation and high performance for deep learning workload, it is intuitive to adapt TVM to Sunway processor. However, the unique compilation environment and architecture features prevent a naive adoption of TVM to Sunway. Firstly, TVM relies on dynamic link libraries to generate executables on different hardware devices, which is not supported on Sunway. In addition, its code organization fails to recognize the different compilation targets for MPE and CPEs, and thus incapable of managing the function calls between MPE and CPEs. Secondly, the memory capacity of each CG on Sunway is quite limited. During the deep learning computation, large memory occupancy is required to store the intermediate data as well as the weight parameters. How to allocate the memory space efficiently and leverage the unique architecture features such as DMA for high bandwidth data transfer is important to generate code with high performance. Thirdly, each CPE within a CG contains a 64KB LDM that can be used to buffer data with explicit software management. How to leverage the limited LDM on each CPE with improved data locality is critical for realizing the performance advantage of Sunway processor during code generation. 

To address the above challenges, we propose \textit{swTVM}, a deep learning compiler tailored for the unique compilation environment and architecture features on Sunway processor. In \textit{swTVM}, we provide ahead-of-time (AOT) code generation that manages the function calls as well as compilation for MPE and CPE explicitly. In addition, we apply several optimizations to the tensor operations so that the architecture features such as DMA and LDM are better utilized during code generation. To the best of our knowledge, this is the first work to implement an end-to-end deep learning compiler on Sunway processor. Specifically, this paper makes the following contributions:

\begin{itemize}
\item We implement the ahead-of-time (AOT) code generation, that produces different compilation targets for MPE and CPE as well as manages the function calls between MPE and CPE efficiently. In addition, we manage the intermediate memory space for each tensor operation globally, which avoids the overhead of frequent memory allocation during computation.

\item We apply several optimizations to the tensor operations regarding the unique architecture features on Sunway. Specifically, we propose a DMA control interface that manipulates the DMA data transfers for each tensor during computation. In addition, we design a LDM management mechanism that buffers the tensor data as much as possible to reduce the latency for accessing memory. Moreover, the DMA instructions are automatically inserted during code generation to improve the accessibility of the buffered data.

\item We propose \textit{swTVM} that implements AOT code generation and architecture specific optimizations on top of TVM, which offers the high performance of Sunway processor to the deep learning community through automatic compilation. The evaluation results show that \textit{swTVM} achieves 1.79$\times$ speedup on average for representative models compared to the state-of-the-art deep learning framework.
\end{itemize}

The rest of this paper is organized as follows. In Section~\ref{sec:background}, we present the background of the deep learning compiler and Sunway processor. Section~\ref{sec:overview} presents the design overview of \textit{swTVM}. Section~\ref{sec:aot} and Section~\ref{sec:tensor} describe the details of code generation in AOT mode and optimizations for tensor operations on Sunway. Section~\ref{sec:evaluation} presents the evaluation results of \textit{swTVM} compared to \textit{swCaffe}. Section~\ref{sec:relatedwork} presents the related work, and section~\ref{sec:conclusion} concludes this paper.

%% file: background.tex
\section{Background}
\label{sec:background}

\subsection{Sunway Processor}
\label{subsec:sunway}

Each Sunway SW26010 processor has four CGs, where each CG contains 1 MPE and 64 CPEs.
The executables on Sunway are generated through cross-compilation on x86 processor using customized compiler. Due to the limitation of the customized operating system on Sunway, it does not support dynamic linked libraries. Instead, the executables are generated with libraries statically linked. Moreover, the codes running on MPE and CPEs are compiled as different compilation targets (using compilation flags of \textit{-host}) and \textit{-slave}, respectively).

As for memory hierarchy, each CPE has 16KB L1 instruction cache and 64KB local device memory (LDM). The LDM is commonly used as a programmable buffer with explicit software control. There are two ways to access main memory on Sunway. The first one is to use DMA, which prefers large and continuous data access. The other one is to use global load/store (Gload/Gstore) instruction, which prefers small and random data access compared to the DMA.

Two parallel programming models are supported on Sunway to exploit the massive parallelism of the CPEs, including OpenACC and Athread. OpenACC is more programmer friendly, with which programmers can utilize CPEs without knowing about the underlying architecture details. While with Athread, programmers can buffer the data in LDM, which provides the opportunity to reduce the accesses to main memory through explicit control. In this paper, we generate Athread codes on Sunway for better performance.

Although the LDM of CPE sounds similar to the shared memory on GPU, their design philosophies are quite different. GPU adopts SIMT parallelism that accesses the shared memory through concurrent threads within a warp. The GPU program achieves better performance if threads within a warp access a continuous memory region at the same time. However, on Sunway the CPEs access the memory and buffer the data in LDM independently. Therefore, without careful management, severe contention on memory bandwidth would occur and thus degrade the performance significantly. In addition, when buffering large continuous data block to LDM, the DMA data transfer can be utilized for higher memory bandwidth.

\subsection{Automated Compilation for Deep Learning}
\label{subsec:dlcompiler}

There are increasing demands of deploying emerging deep learning models to various hardware devices, so that enormous engineering efforts are required to match the algorithms with the hardware efficiently. Currently, the performance of the deep learning models mainly depends on the computation library, such as cuDNN and MKL provided by hardware vendors. However, it is unsustainable to perform labor intensive performance tuning to match various hardware as new algorithms are arising rapidly. The deep learning compiler provides a way to build an efficient mapping between new algorithms and various hardware targets, and thus improves the portability of the deep learning models. 

Despite different implementation approaches adopted by different deep learning compilers, their design philosophies (e.g., two level optimization) are somehow converging~\cite{dlc-survey}. Therefore, we take TVM for illustration. TVM uses the idea of two-level optimization, including graph level and operator level. On graph level, it converts the deep learning models to the computation graph, and then applies optimizations such as operator fusion and data layout transformation. On operator level, it optimizes the code generation targeting specific hardware through loop optimization (e.g., loop tiling and loop unrolling). However, adapting existing deep learning compiler to Sunway processors introduces several challenges to be addressed, due to the unique compilation environment and architecture features of Sunway. 


\subsection{Challenges for DL compilation on Sunway}
\label{subsec:challenges}
The first challenge is that Sunway processor relies on cross-compilation to generate executables and does not support dynamic linked libraries. It prohibits naive adaption of existing deep learning compiler such as TVM to Sunway. Therefore, code generation in AOT mode needs to be supported in the deep learning compiler so that it can compile the executables with static linked libraries. In addition, an efficient code organization is required with AOT code generation in order to support different compilation targets as well as function calls for MPE and CPEs. 
Moreover, the memory capacity of a CG is quite limited compared to the large volume of data generated during the tensor operation. To avoid the overhead of frequent memory allocation during computation, the memory needs to be managed globally in AOT code generation.

The second challenge is to optimize the generated code regarding the unique architecture features of Sunway. Summarizing from existing researches~\cite{lin2017scalable,liu2018towards,li2018multi,swsptrsv} and the benchmarking~\cite{benchmarking-sw}, the key to achieving high performance on Sunway is to \textit{1)} fully utilize the computing resources of CPEs for massive parallelism, and \textit{2)} leverage the LDM of each CPE to alleviate the bottleneck of memory access. Therefore, when the deep learning compiler optimizes the generated codes, the three rules need to be followed: \textit{1)} use the DMA as much as possible when accessing main memory. The DMA requires accessing large and continuous data block, which provides higher memory bandwidth; \textit{2)} leverage the LDM to buffer as much data as possible during the computation. The LDM reduces the latency to access main memory; \textit{3)} minimize the frequency of memory access as much as possible. The computation should exhibit better data locality and re-accessibility after each memory access.

In sum, implementing an end-to-end deep learning compiler requires both adaptions to the compilation environment on Sunway and optimizations targeting the architecture features to improve the performance of generated codes.

%% file: overview.tex
\section{Design Overview}
\label{sec:overview}

\begin{figure}
	\centering
	\includegraphics[width=\linewidth]{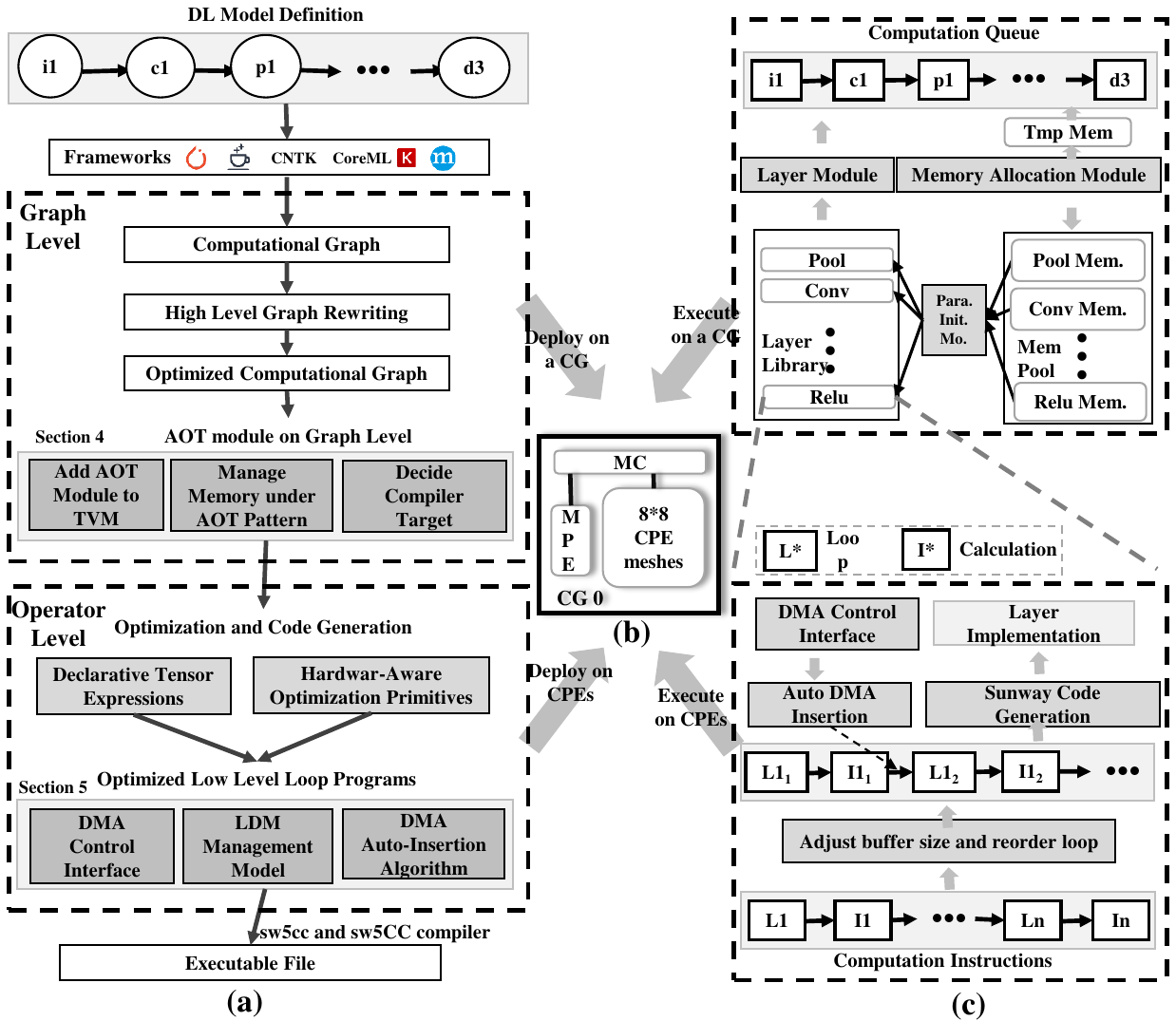}
	\caption{(a) The design overview of \textit{swTVM}, (b) the Sunway architecture and (c) the automatic code generation of deep learning models on MPE and CPEs.}
	\label{fig:main}
\end{figure}

To address the challenges described in Section~\ref{subsec:challenges}, we propose \textit{swTVM} for the Sunway many-core processor. In \textit{swTVM}, we implement the AOT code generation as an extension to TVM, and manage the code organization for MPE and CPEs respectively. In addition, we manipulate the memory allocation of the tensor operation globally. The grey components in Figure~\ref{fig:main}(a) show the contribution of our work. We produce C source codes in AOT mode, which are then compiled by Sunway native compiler in order to generate the executable. The advantage of AOT code generation is that the memory allocation for each layer is determined based on the input and output of each layer before the actual computation, which avoids frequent memory allocation during the computation and thus eliminates the overhead of operations related to memory allocation.

%

The MPE codes generated in AOT mode are primarily responsible for calling each layer according to the topology of the deep learning models, whereas the CPE codes are responsible for the specific computation of operators defined by the layers. The codes generated for a Sunway CG consist of three parts: \textit{layer module}, \textit{memory allocation module} and \textit{parameter initialization module}. To generate the code, the model definition in Figure~\ref{fig:main}(a) is transformed and stored by layer in the \textit{computation queue} in the upper part of Figure~\ref{fig:main}(c). The layer module invokes the layer implementations from the \textit{layer library}, and the memory allocation module allocates the memory space for each layer within the computation queue. The parameter initialization module is responsible for initializing the parameters of the layer implementations within the layer library.

To leverage the architecture features on Sunway, we optimize the implementation of each operator, as shown in the bottom part of Figure~\ref{fig:main}(c). Specifically, we design a DMA control interface, which provides the DMA schedule primitives for the layer library. In addition, since the LDM on each CPE is only 64KB which cannot store the entire tensors, we design a LDM management mechanism to control the amount of tensor data to be buffered in LDM automatically. It can also adjust the buffer size and reorder the computation loops according to the configuration of each layer. Moreover, to improve the locality of the buffered data, we design an algorithm to insert DMA instructions into the appropriate locations of the generated code automatically. The \textit{code generation module} then generates code with Sunway syntax and provides the layer implementation into the layer library, which is utilized by the layer module to fulfill the layers in the computation queue.

Note that, although \textit{swTVM} is targeting the Sunway many-core processor, the approaches are also valuable when building end-to-end deep learning compilers for other emerging processors with cache-less design.

%% file: aot.tex
\section{AOT Code Generation}
\label{sec:aot}

\begin{figure}
    \centering
    \includegraphics[width=\linewidth]{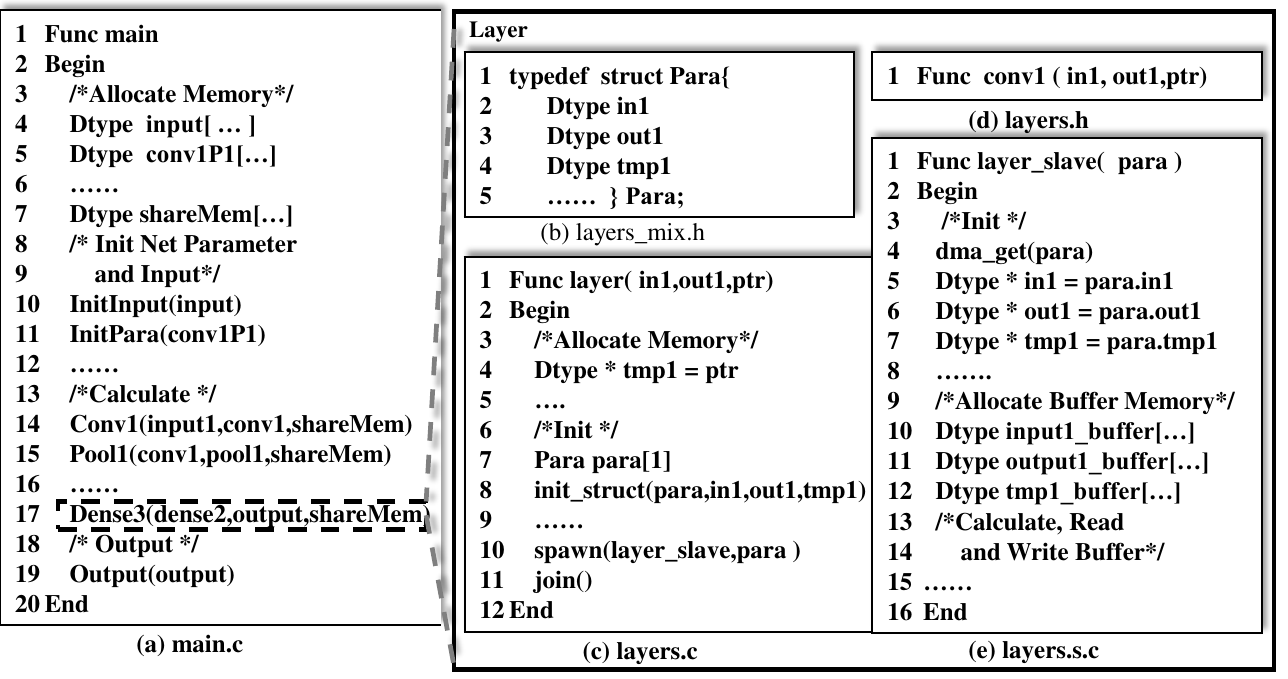}
    \caption{AOT code generation on Sunway processor.}
    \label{fig:aot}
\end{figure}

To implement AOT code generation, we should consider the implementation of each layer and the approach to convert the deep learning model topology into the function calls of layers with the dependencies satisfied. \textit{swTVM} transforms the model topology into the actual implementation on Sunway processor, as shown in Figure~\ref{fig:aot}.
After code generation, the implementation contains a series of operations such as memory allocation, parameter initialization, and function calls in the main function (e.g., \textit{Func main}).

Since the MPE are cores with complete functionality, the generated codes can run on MPE directly. Whereas for CPEs, we need to generate separate files for compiling, as shown in Figure~\ref{fig:aot}. We use a \texttt{struct} to accept multiple parameters in the CPE function (Figure~\ref{fig:aot}). In order to remove the dependency on the struct definition from the interface when calling the layer, we encapsulate CPE functions with another interface that renders the layer function calls as ordinary function calls (Figure~\ref{fig:aot}(d)). The encapsulating interface is also useful when handling the memory allocation of the intermediate data for complex layers. The encapsulated function is organized in a separate file (Figure~\ref{fig:aot}(d)) with MPE as its compilation target. The parameters stored in the struct file is only visible to the files containing the CPE function and encapsulated CPE function. We achieve the AOT code generation for each layer by organizing the code of each layer into the above three files in addition to a header file (Figure~\ref{fig:aot}(c)) for the encapsulated CPE function.

\subsection{Managing Memory Allocation}
\label{subsec:memalloc}
The memory allocation on both main memory and LDM for input/output data as well as temporal data of each layer needs to be managed explicitly. The memory allocated for input/output data includes intermediate data generated between layers, and weight parameters that cannot be released or overwrote during computation. Once completing one layer, each operator stores its result into main memory and then used by other operators. Since this data is stored in the main memory, the memory space is allocated and freed by MPE, as shown in Figure~\ref{fig:aot}(a) (line 3-7).

Complex operators usually generate temporal data. The data is never re-used and thus can be freed once the computation completes. Because the temporal data is usually larger than the capacity of LDM (i.e., 64KB), it is also stored in the main memory, whose allocation and deallocation are controlled in the main function. When an operator is invoked, it uses a portion of the memory space that has already been allocated in the main function, which reduces the overhead for allocation and deallocation for each operator. The memory space for temporal data is allocated by MPE and used by CPEs. The implementation details are listed in \textit{Func main} for MPE and \textit{Func layer\_slave} (e.g., in layers.s.c file) for CPEs in Figure~\ref{fig:aot}. The LDM utilization in \textit{Func Layer\_slave} is described in Section~\ref{sec:tensor}.

\subsection{Managing Function Call}
\label{subsec:func}
As shown in Figure~\ref{fig:main}(a), the implementation of \textit{swTVM} is organized into three levels, which first transforms the topology of a deep learning model into computation graph, and then applies a serial of optimizations at graph level, and eventually implements the computation on specific hardware at operator level. In AOT code generation, the \textit{Func main} in Figure~\ref{fig:aot}(a) is responsible for maintaining the dependency of function calls in the computation graph, whereas \textit{Fun layer\_slave} in Figure~\ref{fig:aot}(e) implements each operator. Note that the \textit{Func layer} in Figure~\ref{fig:aot}(c) is the interface that connects \textit{Fun layer\_slave} and \textit{Func main}, and fulfills the function call of each operator in the computation graph.

In addition, function calls for architecture specific codes can also be organized into three levels, including function call on MPE, function call on CPEs and function call from MPE to CPEs. These three levels correspond to the operators at graph level, operator level, and from graph level to operator level in \textit{swTVM}. 
The graph level generates \textit{Func Main}, which runs on MPE. And the \textit{Func layer} is the implementation of the function call from graph level to operator level, which is invoked by \textit{Func main} on MPE and then invokes the \textit{Func layer\_slave} on CPEs. \textit{Func layer\_slave} implements the computation performed at operator level.

With such design, \textit{swTVM} can organize the AOT code generation and Sunway architecture optimizations through layered function calls, rather than relying on sophisticated low-level implementation details. In addition, through managing the dependencies of function calls, \textit{swTVM} is able to generate codes for MPE and CPEs as different compilation targets.

\subsection{Implementation Details}
\label{subsec:details}
The \textit{Func main} shown in Figure~\ref{fig:aot}(a) consists of four stages, including memory allocation stage, parameter/input initialization stage, computation stage and output stage. During memory allocation stage, in addition to memory for the parameter and input/output of the deep learning model, temporal memory is also allocated for each layer, the size of which satisfies the maximum memory usage of each layer. The dependency across all layers is analyzed to decide  the order of function calls. Each function in the computation stage corresponds to one or more layers in the model topology.

The implementation of each operator consists of \textit{Func layer} on MPE, \textit{Func layer\_slave} on CPEs and parameter structure \textit{Para}. \textit{Func layer} is further divided into three parts, such as the memory allocation for temporal space, parameter initialization, and computation. The memory allocation for temporal space is only required for the layer that combines multiple sub-operators such as convolution and pooling. For such layers, the input of one sub-operator depends on the intermediate results from the previous sub-operator. Considering the overhead of frequent memory allocation, we allocate temporal memory space in the main function and share it across operators.

The format for calling the function on CPEs is to use the function name and parameter struct, as shown in Figure~\ref{fig:aot}(c) (line 10). \textit{Para} is the parameter struct that is only visible to corresponding \textit{layer.c} and \textit{layer.slave.c} files. \textit{Func layer\_slave} consists of parameter parsing, LDM allocation, and computation. At the beginning of the function, the tensors are loaded from memory and then buffered in LDM. The LDM space is allocated through static arrays to buffer the tensor. The main memory is accessed through DMA instructions, which can be overlapped with the computation for efficiency, and the details are described in Section~\ref{sec:tensor}.

\subsection{Invoking Optimized Kernel Libraries}
Since there are several optimized libraries available on Sunway processor for accelerating matrix multiplication and convolution computation, such as \textit{xMath}, \textit{swGEMM} and \textit{swDNN}~\cite{fang2017swdnn}. \textit{swTVM} provides optional approach to easily integrate these libraries for better code generation, which is implemented in the following two stages. In schedule mapping stage, \textit{swTVM} uses the intrinsic APIs to generate function calls to external libraries, and bypasses them to the code generation stage. 
Considering performance variation of different libraries across different tensor operations, \textit{swTVM} invokes the libraries with optimal performance. 
For example, it invokes \textit{xMath} and \textit{swGEMM} for accelerating standard convolution and depthwise convolution operators respectively.
In code generation stage, \textit{swTVM} identifies the invoked libraries and automatically adds the relevant header files and parameters to generate the canonical C codes. Besides, it adds the corresponding flags to the compilation configurations (e.g., Makefile).

%% file: tensor.tex
\section{Optimizing Tensor Operation}
\label{sec:tensor}

\subsection{DMA Control Interface}
\label{subsec:dmainter}
\begin{figure}
    \centering
    \includegraphics[width=\linewidth]{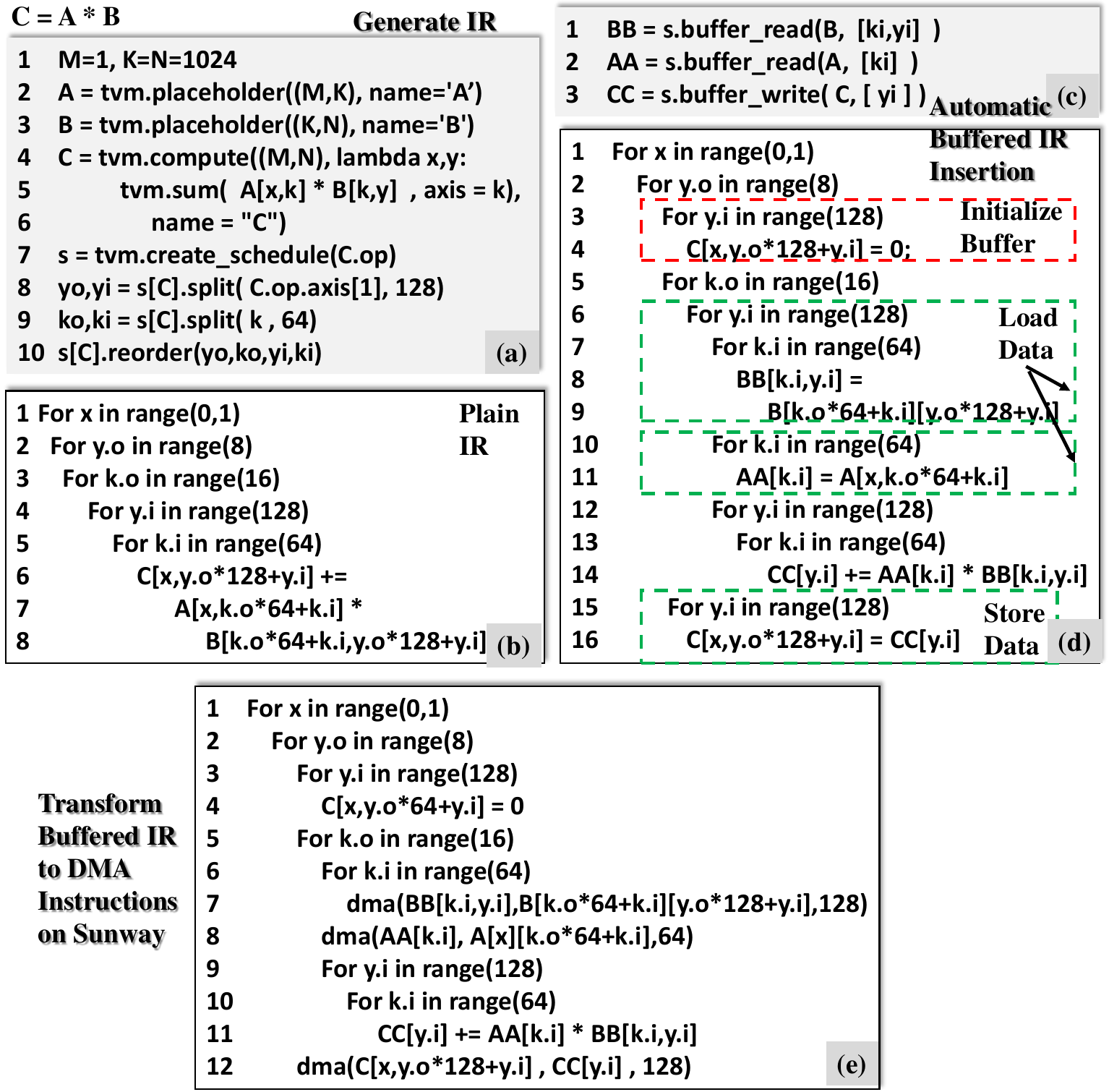}
    \caption{An example of matrix multiplication implementation generated by \textit{swTVM} with optimizations targeting Sunway.} 
    \label{fig:dma}
\end{figure}

An efficient DMA control interface plays an important role in \textit{swTVM} to generate high-performance implementations of deep learning models on Sunway. 
In \textit{swTVM}, the DMA control interface provides schedule primitives to control DMA in order to manage the data access efficiently. Figure~\ref{fig:dma} shows an example to control the tensor data access in matrix multiplication through the DMA control interface. Figure~\ref{fig:dma}(a) shows the computation definition in \textit{swTVM}, and Figure~\ref{fig:dma}(b) shows the plain IR generated by \textit{swTVM}, which is the same as original TVM. 
The \textit{split} primitive splits the loop iterator into two parts (line 8-9 of Figure~\ref{fig:dma}(a)). We call the outer part as parallel iterator and inner part as buffer iterator. And \textit{swTVM} uses the parallel iterators to assign computation to CPEs for parallelization and buffer iterators for DMA data transfer.

\textit{swTVM} can also buffer data along multiple dimensions. In Figure~\ref{fig:dma}(c), tensor \textit{B} is buffered along two dimensions (line 1). This allows fast access to the value along these two dimensions of tensor \textit{B} when calculating the sub-region of tensor \textit{C}. 
Additionally, \textit{swTVM} can specify which tensor to be buffered and the region of the tensor to be buffered during code generation. 
To buffer partial of the tensor along one dimension, \textit{split}, \textit{buffer\_read}, and \textit{buffer\_write} primitives are applied in sequence to split the dimension and buffer the corresponding data. After invoking the above primitives, \textit{Load Data} region (line 6-11 in Figure~\ref{fig:dma}(d)) generates the IR code of the read buffer for tensor \textit{B} and \textit{A}, whereas \textit{Store Data} region (line 15-16 in Figure~\ref{fig:dma}(d)) generates the IR code of write buffer for tensor \textit{C}. The generated IR is then translated to DMA instructions during code generation.

Buffering data along multiple dimensions also occurs in convolution operator. The convolution operation is the computation among high-dimension tensors, where certain dimensions of the tensor may be quite small
If only buffering data along only one dimension, the LDM space is not fully utilized. In such a case, buffering the tensor data along multiple dimensions improve the LDM utilization. When buffering, we satisfy the data access of the outer loop with high priority, which improves the locality of buffered data.

In complex layer such as convolution, the subscript to access the tensor data along one dimension is determined by multiple loop iterators. To handle such case, the DMA control interface accepts multiple loop iterators and allows the user to specify the expression on calculating the subscript based on these loop iterators, which determines the range of each dimension to be buffered. One such example is shown in the expression $yy \times stride+ry$. The DMA control interface also supports expression inferring, which accepts the subscript expression and analyzes the correlation between the loop iterators and tensor dimensions automatically.


\subsection{LDM Management Mechanism}
\label{sunsec:ldm}
\begin{figure}
    \centering
    \begin{minipage}[t]{0.485\linewidth}
        \centering
        \includegraphics[width=\linewidth]{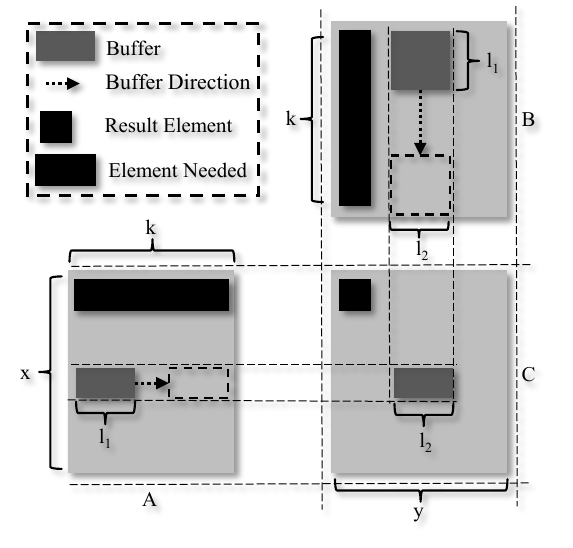}
        \caption{Buffer size dependency of matrix \textit{A}, \textit{B}, and \textit{C} within matrix multiplication.} 
        \label{fig:need_ldm}
    \end{minipage}
    \hfill
    \begin{minipage}[t]{0.485\linewidth}
        \centering
        \includegraphics[width=\linewidth]{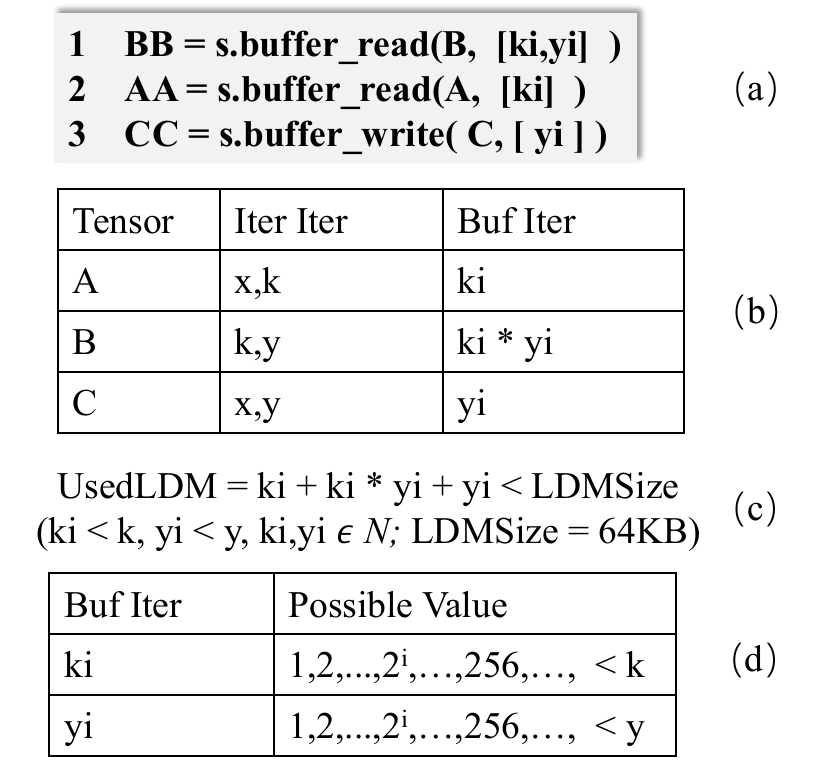}
        \caption{ Procedure of calculating the buffer size on the matrix multiplication.} 
        \label{fig:manage_ldm}
    \end{minipage}
\end{figure}

To better control the data buffering in LDM, we design the LDM management mechanism, which determines the buffer size and the dimensions of tensor to be buffered. In addition, it reorders the computation loops to improve the locality of the buffered data.

\subsubsection{Determining the Buffer Size}
Due to the limited LDM space, the difficulty to determine the buffer size of each tensor is to identify the dependencies, which means the buffer size of one tensor can affect the buffer size of another. Figure~\ref{fig:need_ldm} shows an example of buffer size dependency within the matrix multiplication ($A\times B = C$). The dimensions of  matrix \textit{A}, \textit{B} and \textit{C} are \textit{(x,k)}, \textit{(k,y)}, and \textit{(x,y)}, respectively. When the buffer size of matrix \textit{C} and matrix \textit{A} is $l_{2}$ and $l_{1}$ respectively along the same dimension, the buffer size of matrix \textit{B} is $l_{1}$ and $l_{2}$ along \textit{k} and \textit{y} dimension.

Figure~\ref{fig:manage_ldm} shows the procedure of calculating the buffer size on the matrix multiplication and the possible size of each buffer. When determining the buffer dimension of each tensor using \textit{buffer\_read} and \textit{buffer\_write} in Figure~\ref{fig:manage_ldm}(a), \textit{swTVM} constructs a table that describes the buffer iterators of each tensor as shown in Figure~\ref{fig:manage_ldm}(b). The sum of the buffer size from all buffer iterators of each tensor can be expressed in an equation, as shown in Figure~\ref{fig:manage_ldm}(c). Then the buffer size of each tensor can be determined by choosing a possible value that satisfies the above equation. We limit the possible values to the power of two for better performance on Sunway, as shown in Figure~\ref{fig:manage_ldm}(d). We use a greedy algorithm to search for the minimum possible value.

\textbf{Constraint 1 -} When determining the buffer size of each tensor, the following constraints should be satisfied.
\begin{itemize}
    \item The buffer size of a buffer iterator cannot be larger than the original dimension.
    \item The sum of the buffer size of all tensors cannot be larger than the LDM size.
    \item When the buffer size in the lowest dimension is one, no buffer is allocated for the tensor, which means that the data is directly accessed from main memory.
\end{itemize}

\textbf{Strategy 1 -} \textit{swTVM} uses an approximate algorithm (Algorithm~\ref{alg:ldm} in Appendix) to managing data buffer in LDM, which ensures an acceptable search time for an optimal solution. The algorithm consists of two parts, the initial part to allocate a pre-defined LDM memory space and expanding part to maximize the LDM utilization. In Algorithm \ref{alg:ldm}, the iterators can be classified into three types:

\begin{itemize}
    \item \textit{sizeiter}: determines the buffer size and is the index of the lowest dimension of the tensor that can be expanded;
    \item \textit{numiter}: determines the number of DMA instructions and is the index of the dimension (except the lowest dimension) of the tensor;
    \item \textit{compiter}: is the iterator that satisfies the conditions of both \textit{sizeiter} and \textit{numiter}.
\end{itemize}

At the beginning of the algorithm, the sequence of the iterators is reordered. For \textit{compiters}, it is reordered by the ascending order of the affected number of tensors. Whereas for \textit{sizeiters}, it is reordered by the ascending order of the buffer size (line \ref{alg:ldm:sort:begin}-\ref{alg:ldm:sort:end}). After that, the buffer iterator for each loop iterator is initialized to a pre-defined size across each tensor (line \ref{alg:ldm:init:begin}-\ref{alg:ldm:init:end}). The buffer size is set to the minimum between the loop range and \textit{InitValue}.
The number \textit{InitValue} is chosen based on empirical study that reading \textit{InitValue} floats per memory access achieves good bandwidth, and \textit{InitValue} is set to 64 on Sunway (line \ref{alg:ldm:64:begin}-\ref{alg:ldm:64:end}). Then, the algorithm checks if the buffer size is larger than the size of LDM. If so, the amount of data to be transferred for current buffer iterators or even the previous buffered iterators needs to be reduced to fit in the limited size of LDM (line \ref{alg:ldm:check:begin}-\ref{alg:ldm:check:end}).

During the initialization, the algorithm invokes the \textit{UPDATE} function if the range of the buffer iterator equals to the range of the loop iterator. 
When the dimension of the tensor to be buffered is no longer associated with any iterators, the higher dimension needs be adjusted to change the buffer size. And the \textit{CLASSIFY} function is invoked to update \textit{numiters}, \textit{sizeiters} and \textit{compiters} (line \ref{alg:ldm:dim:begin1}-\ref{alg:ldm:dim:end1}, \ref{alg:ldm:dim:begin3}-\ref{alg:ldm:dim:end3}, \ref{alg:ldm:dim:begin2}-\ref{alg:ldm:dim:end2}). After the initialization, if the LDM still has free space, the buffer size of each iterator is expanded to improve the LDM utilization. We use a greedy algorithm to load as much data into LDM as possible. The algorithm terminates when the LDM usage reaches the maximum size (line \ref{alg:ldm:extent:begin}-\ref{alg:ldm:extent:end}).

We take the matrix multiplication in Figure~\ref{fig:dma} to illustrate the process of the algorithm, where \textit{x}, \textit{y} and \textit{k} is \textit{numiter}, \textit{sizeiter}, and \textit{compiter} respectively. We set the buffer size of \textit{y} to 64 and ensure our buffer size not exceeding the LDM capacity. Then, we set \textit{k} to 64 that leads to the LDM usage of 16.5KB. Since there is no \textit{numiter} satisfying the condition of \textit{UPDATE}, the algorithm enters the expanding part. When \textit{y} is set to 128, the LDM usage increases to 32.75KB. Continuing to expand \textit{k} to 128, the buffer size reaches 65KB, which is larger than the LDM capacity (64KB). Therefore, \textit{x} = 1, \textit{y} = 128, and \textit{k} = 64 are chosen as the buffer sizes. 


\subsubsection{Loop Reordering}
After initializing the buffer size for each tensor, the loop order is adjusted to improve the locality of the buffered data. 

\textbf{Constraint 2 -} To ensure the correctness after loop reordering, the following constraints need to be satisfied.
\begin{itemize}
    \item The buffer iterator cannot be ahead of the parallel iterator, both of which are split from the same iterator;
    \item The parallel iterator of the output tensor must be at the outermost loop to prevent write conflict;
    \item The buffer cannot be ahead of other iterator associated with the same tensor;
    \item The child iterator split from the parent iterator inherits its parent's order.
\end{itemize}

\textbf{Strategy 2 -} Under the above constraints, we reorder the loop iterators that are not associated with the tensor and insert the DMA instruction into the suitable location to avoid unnecessary DMA transfers. For the conflicting DMA instructions, the loops are reordered, and the loop order with the least number of DMA instructions is chosen, as shown in Algorithm~\ref{alg:order} in Appendix. First, all buffered iterators are moved to the innermost loop. And then, the order of non-buffered iterators are determined. The buffered iterators with locations undecided are inserted to the current loop with the number of DMA instructions for all tensors evaluated. The iterators with the least number of DMA instructions is chosen as the loop iterators for current loop (line \ref{alg:order:select:begin}-\ref{alg:order:select:end}). The above process is repeated until all iterators are evaluated, which derives the final loop order (line \ref{alg:order:loop:begin}-\ref{alg:order:loop:end}). The time complexity of Algorithm~\ref{alg:order} is also within polynomial time.

We take the matrix multiplication in Figure~\ref{fig:dma} to illustrate the loop reordering. The iterators for which the order to be decided is \textit{x}, \textit{yo} and \textit{ko}. The least number of DMA instructions for \textit{x}, \textit{yo} and \textit{ko} is $256+128$, $256+16$ and $256+8$ respectively. Therefore, \textit{ko} is chosen first. And then, the least number of DMA instructions for \textit{x} and \textit{yo} are both 8. Therefore, the original loop order is unchanged. After that, the final loop order for \textit{x}, \textit{yo} and \textit{ko} is determined.

\subsection{DMA Auto-Insertion Algorithm}
\label{subsec:dmainsert}
\begin{figure}
    \centering
    \includegraphics[width=\linewidth]{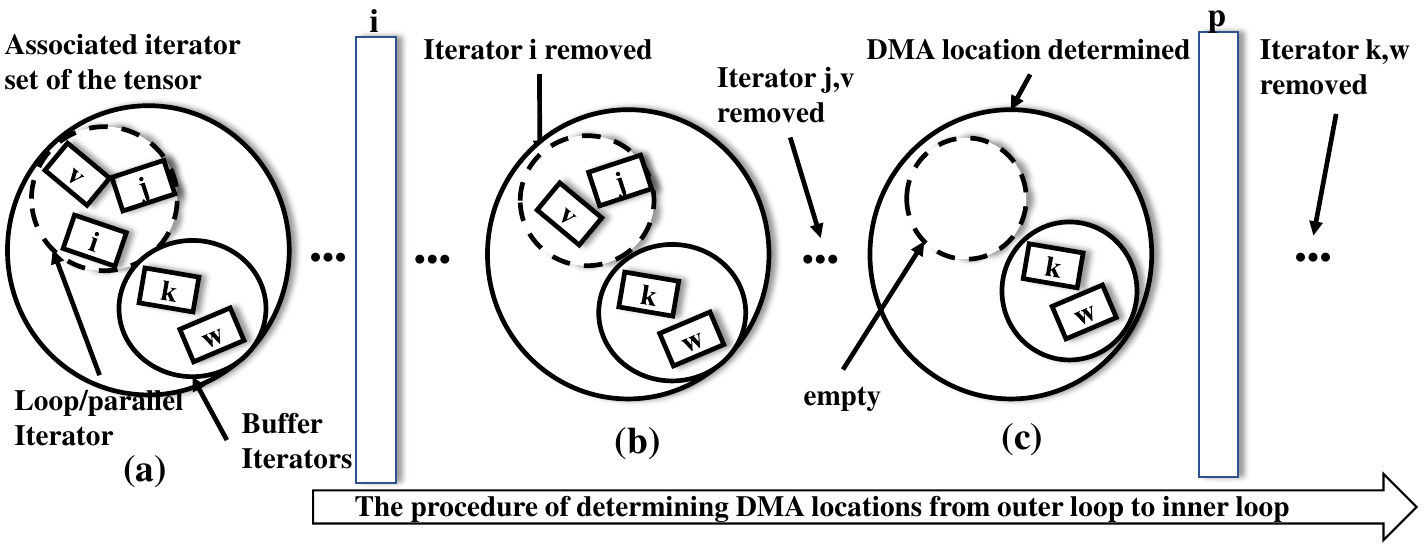}
    \caption{The illustration of DMA auto-insertion algorithm. (a) the initial states of iterators, (b) iterator \textit{i} to be removed and (c) the DMA locations are determined for the tensor.} 
    \label{fig:auto_insert}
\end{figure}
With the DMA control interface and LDM management mechanism available, we propose an algorithm to implement the auto-insertion of DMA instructions during the code generation. 
The DMA auto-insertion algorithm 
consists of three parts as following.

\textbf{Determining the buffer size and the starting memory location}. First, the buffer dimension is split into two parts, which makes the range of the inner loop within the buffer size. When the subscript of the dimension is correlated with only one loop iterator, the starting memory location of the buffer is calculated by setting the loop iterator of the inner loop to 0, whereas the buffer size is the range of the inner loop. All the buffer operations in Figure~\ref{fig:dma}(c) belong to the above case. However, for complex operators such as stride convolution, the subscript of one dimension of the tensor is always correlated with several loop iterators. To obtain the starting memory location of the buffer, all loop iterators are set to zero and calculated in the subscript expression. The size of the buffer is the difference between the result of the subscript expression with all iterators set to their maximum value and the starting memory location.

\textbf{Determining the locations of DMA instruction insertion}. 
Figure~\ref{fig:auto_insert} illustrates the process of determining the DMA insertion location for a tensor. At the beginning, we have the iterators which are associated with the tensor. Figure~\ref{fig:auto_insert}(a) shows the initial states of the iterators. The tensor has loop iterators such as \textit{v}, \textit{j} and \textit{i}, and buffer iterators such as \textit{k} and \textit{w}. Then we iterate through the outer loops to inner loops. If the iterator of the current loop is not within the associated iterator set of the tensor, then the algorithm proceeds to the next loop. If the current iterator belongs to the set but does not belong to the buffer iterators, then the iterator is removed from the associated iterator set, which indicates the iterator is determined. In Figure~\ref{fig:auto_insert}(b), the iterator \textit{i} is removed from the associated iterator set since it satisfies the above condition. When all iterators except the buffer iterators are removed from the associated iterator set, as shown in Figure~\ref{fig:auto_insert}(c), the locations to insert DMA instructions are determined. The above procedure is repeated for all tensors to determine the locations of DMA instruction insertion correspondingly.

\textbf{Generating code with Sunway DMA syntax}. Figure~\ref{fig:dma}(e) shows the pseudo-code of the inserted DMA instructions. When generating the code, the DMA instructions whose memory address and LDM buffer address are continuous, are combined to reduce the number of DMA instructions.

\subsection{Parallelism}
\label{subsec:tensor:implement}

To achieve better parallel efficiency with CPEs, the load balance and write conflict need to be considered when generating codes. The load balance can be achieved by using the \textit{athread\_parallel} primitive, which splits computation task into sub-tasks along the highest dimension of the tensor. Take the vector multiplication ($v = v1 \times v2$) with parallel implementation as an example. For the vector \textit{v} with dimension size of 1,024, we divide its dimensions into \textit{CoreNum} chunks. As \textit{CoreNum} on Sunway is 64, the size of each chunk is 16. The \textit{\_begin} and \textit{\_end} indicates the range of sub-tasks for each CPE, which is determined by the \textit{id} of CPE and the number of the tasks. The less optimal case happens when the size along the high dimension of the tensor is less than the number of CPEs. Such a case can be solved by using the \textit{fuse} primitive to combine multiple dimensions until the size is large enough. And the write conflict can be avoided by splitting the tasks along the dimension of the tensor to be written.

%% file: evaluation.tex
\section{Evaluation}
\label{sec:evaluation}
\subsection{Experiment Setup}
In this section, we evaluate the performance of the codes generated by \textit{swTVM} on a CG of Sunway processor. 
We compare \textit{swTVM} with \textit{swCaffe}~\cite{li2018swcaffe}, which is the deep learning framework customized for Sunway. 
We present the end-to-end performance and the operator-level performance to demonstrate the efficiency of \textit{swTVM}, and we provide the roofline model analysis to better understand the generated codes. 
Besides, we show the compilation overhead of \textit{swTVM}.

For benchmarks, we select eight representative deep learning models that are the widely-used in inference tasks, as shown in Table~\ref{tab:benchmarks}. 
Notably, \textit{swCaffe} fails to execute ShuffleNet and Bert-base due to unsupported layers (e.g., permute, layernorm, and embedding).
While \textit{swTVM} supports them and generates high-performance codes for them, demonstrating its portability.
Each model under each batch size is executed for 100 times and the average execution time is reported. 

\begin{table}[htbp]
    \centering
    \footnotesize
    \caption{Deep learning models in experiments.}
    \label{tab:benchmarks}
    \begin{tabular}{|c|c|c|c|}
        \hline 
        Model & Task & Batch Size ($bs$) & Input Size \\ \hline
        ResNet18 & Image Classification & 1, 2, 4, 8 & ($bs$,3,224,224) \\ \hline
        ResNet50 & Image Classification & 1, 2, 4, 8 & ($bs$,3,224,224) \\ \hline
        VGG16 & Image Classification & 1, 2, 4, 8 & ($bs$,3,224,224) \\ \hline
        YOLOv3 & Object Detection & 1, 2 ,4 ,8 & ($bs$,3,416,416) \\ \hline
        DCGAN & Image Classification & 1, 2, 4, 8 & ($bs$,100,1,1) \\ \hline
        MobileNet & Image Classification & 1, 2, 4, 8 & ($bs$,3,224,224) \\ \hline
        ShuffleNet & Image Classification & 1, 2, 4, 8 & ($bs$,3,224,224) \\ \hline
        Bert-base & Question Answering & 1, 2, 4, 8 & ($bs$,seqlen=16) \\ \hline
    \end{tabular}
\end{table}  

\textit{swTVM} performs the compilation on the x86 platform. The compilation environment includes gcc/g++ 4.8.5 and Python 3.6.2. The codes generated by \textit{swTVM} is a group of C/C++ files, which are then compiled by Sunway native compilers (sw5cc for C and sw5CC for C++) with \textit{-O3}.
\textit{swCaffe} is configured with the recommended high-performance libraries, including \textit{swDNN} and \textit{xMath}. 
And all optimization macros (e.g., offloading \textit{im2col}, \textit{relu}, \textit{pooling}, \textit{batchnorm} operators to \textit{swDNN}) are enabled.

\subsection{End-to-End Performance}
\label{subsec:e2e}

\begin{figure*}[h]
    \centering
    \begin{subfigure}{0.49\linewidth}
        \centering
        \includegraphics[width=\linewidth]{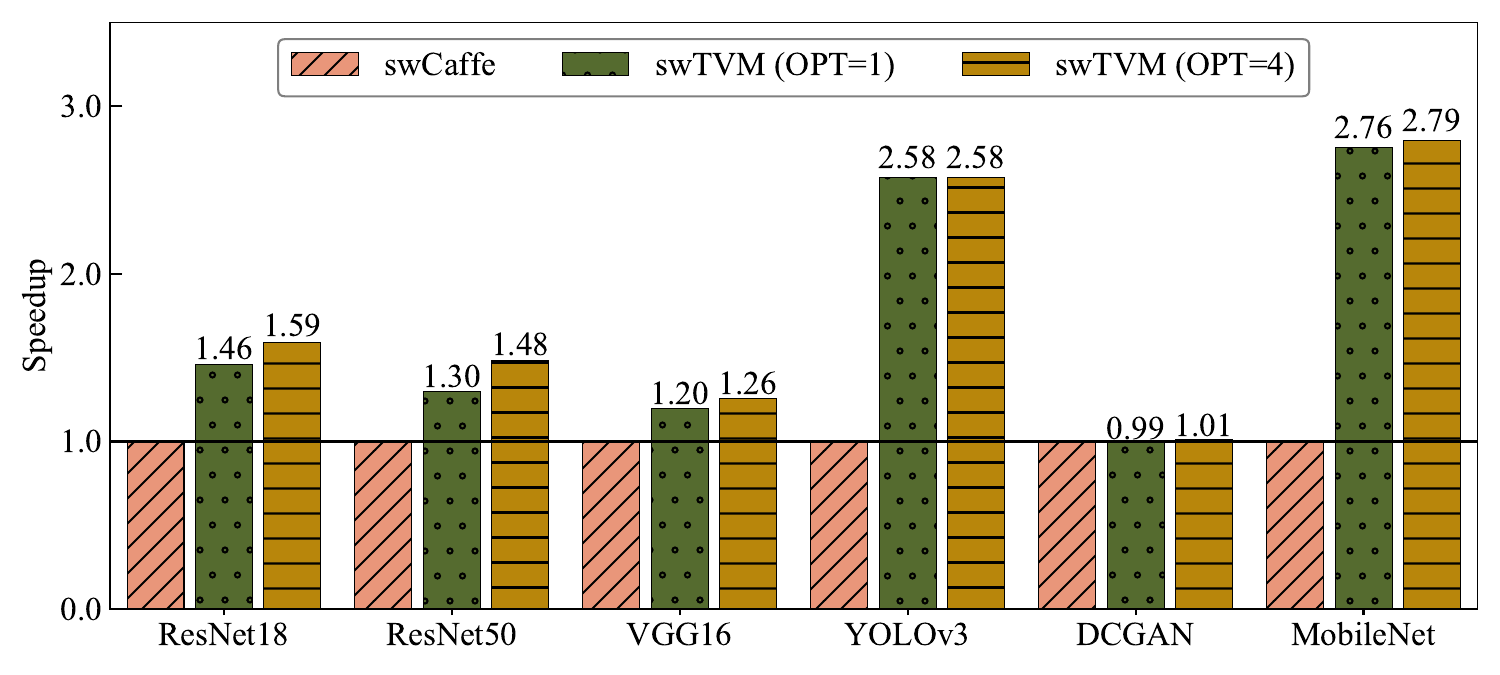}
        \vspace{-0.2in}
        \caption{batch size = 1}
    \end{subfigure}
    \begin{subfigure}{0.49\linewidth}
        \centering
        \includegraphics[width=\linewidth]{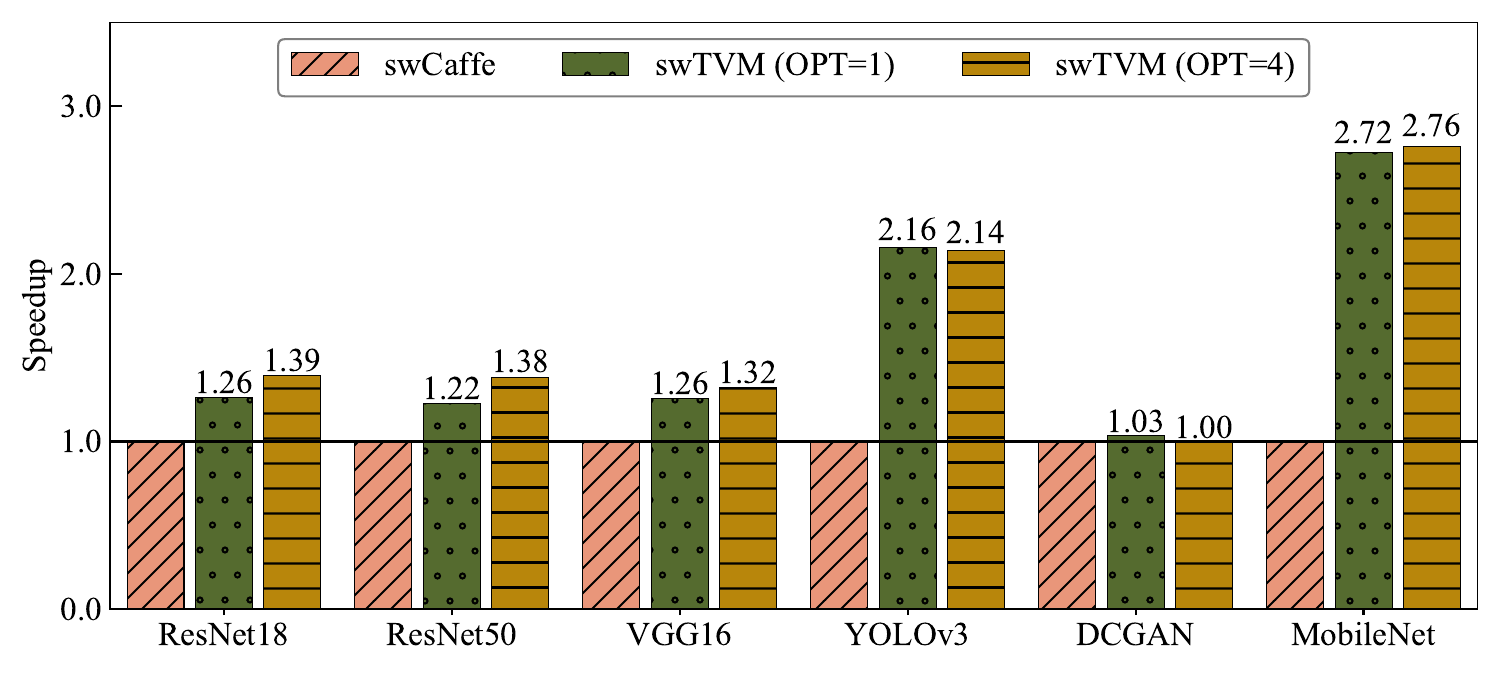}
        \vspace{-0.2in}
        \caption{batch size = 2}
    \end{subfigure}
    \begin{subfigure}{0.49\linewidth}
        \centering
        \includegraphics[width=\linewidth]{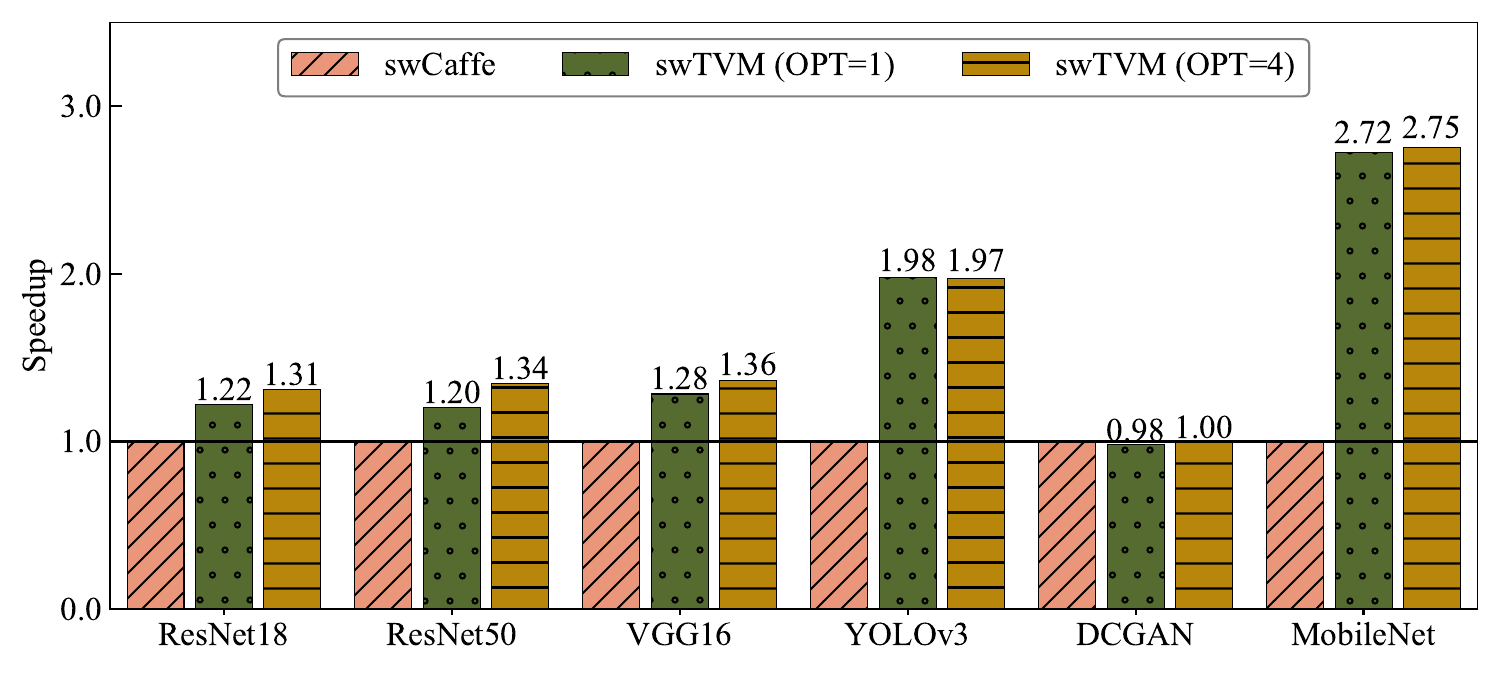}
        \vspace{-0.2in}
        \caption{batch size = 4}
    \end{subfigure}
    \begin{subfigure}{0.49\linewidth}
        \centering
        \includegraphics[width=\linewidth]{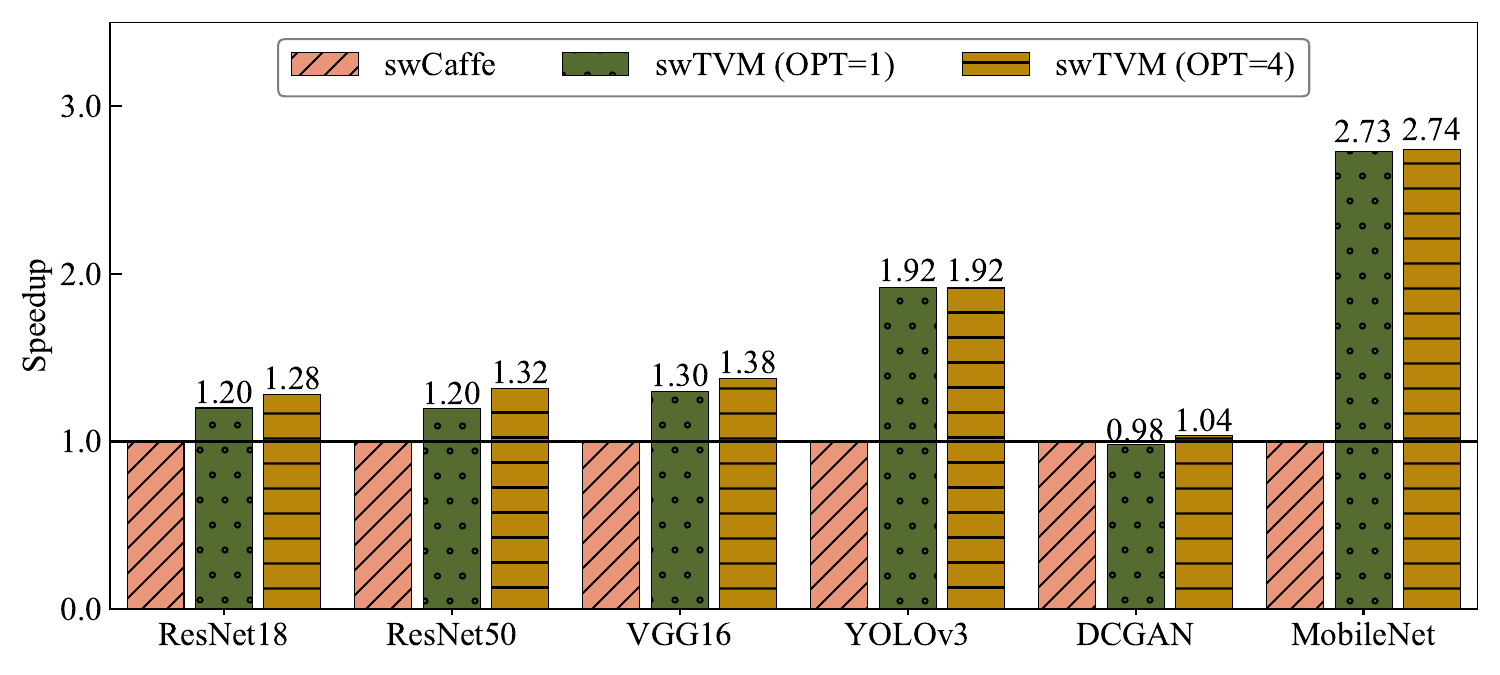}
        \vspace{-0.2in}
        \caption{batch size = 8}
    \end{subfigure}
    \caption{End-to-end performance of \textit{swTVM} with two configuration of graph-level optimization, OPT=1 and OPT=4. The y-axis represents the speedup compared to \textit{swCaffe}.}
    \label{fig:e2e}
\end{figure*}

The end-to-end performance of \textit{swTVM} across all benchmarks is shown in Figure~\ref{fig:e2e}. 
Notably, \textit{swTVM} is configured with two configurations of graph-level optimizations: {OPT=1} enables the basic operator fusion, and {OPT=4} enables all built-in optimization passes of TVM.
\textit{swTVM} under both configurations outperforms \textit{swCaffe} in nearly all benchmarks. 
Specifically, the average speedups of \textit{swTVM} (OPT=1) compared the \textit{swCaffe} baseline under the four batch sizes (i.e., 1, 2, 4, 8) are 1.71$\times$, 1.61$\times$, 1.56$\times$, and 1.55$\times$, respectively. 
And the average speedups of \textit{swTVM} (OPT=4) are 1.79$\times$, 1.66$\times$, 1.62$\times$, and 1.61$\times$. 
This is because \textit{swTVM} exploits the graph-level optimizations standing on the basis of TVM, while \textit{swCaffe} ignores them.
For example, the operator fusion can reduce the number of kernel launches on CPEs, eliminate the corresponding DMA transfer between LDM and main memory, and allow better sharing of the computation. 
With more graph-level optimizations enabled, the performance of \textit{swTVM} (OPT=4) is better than that of \textit{swTVM} (OPT=1).

The acceleration from memory-intensive operators (e.g., batch\_norm) dominates the performance improvement of \textit{swTVM}.
Among the benchmarks, YOLOv3 has the largest batch\_norm operators (with the largest input size), and thus it has higher speedup ration.
With the increasing batch sizes, the speedup of \textit{swTVM} decreases slightly, because of the inefficient batch\_norm implementation of \textit{swCaffe} baseline.
\textit{swCaffe} implements batch\_norm through matrix multiplication which shows non-trivial overhead, therefore, the computation time remains nearly constant even doubling the batch size.
Notably, as for MobileNet, the depthwise\_conv2d operators dominate over 95\% of the computation time and are also highly-optimized by \textit{swTVM}.
Consequently, on MobileNet, \textit{swTVM} achieves the maximum speedup of 2.79$\times$, which is quite stable even with different batch sizes.

\subsection{Operator-Level Performance}
\label{subsec:layer}

\begin{figure}
        \centering
        \includegraphics[width=\linewidth]{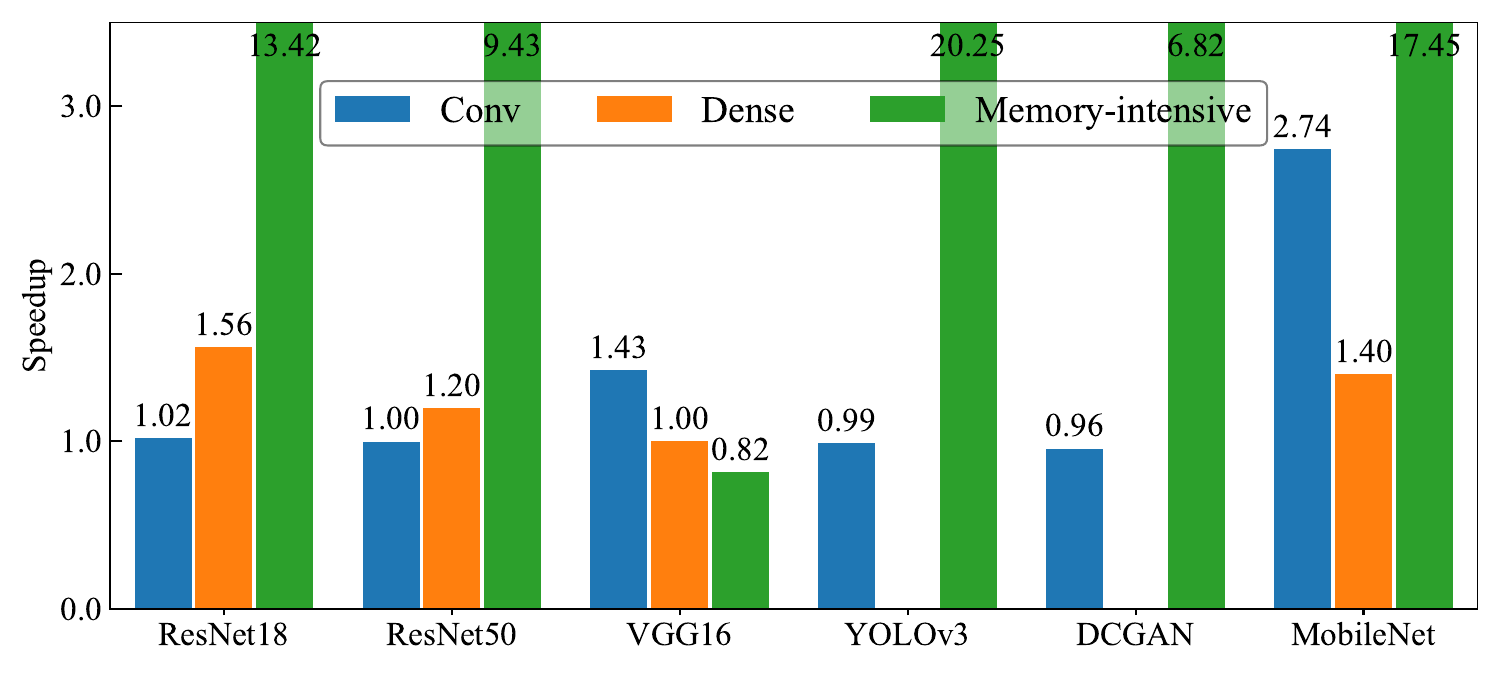}
        \caption{Performance of convolution, dense, and memory-intensive layers of \textit{swTVM} compared to \textit{swCaffe}, when batch size is set to 1.}
        \label{fig:layer}
\end{figure}

In order to evaluate the effectiveness of \textit{swTVM}, we further perform operator-level performance comparison.
We classify the operator into three categories, including convolution, dense, and memory-intensive operators. 
The experiment results is shown in Figure~\ref{fig:layer}. Since YOLOv3 and DCGAN has no dense operator, the corresponding bars are left blank.
On the three categories, \textit{swTVM} achieves 1.36$\times$, 1.29$\times$, and 11.36$\times$ speedups on average compared to \textit{swCaffe}, respectively.
Among them, \textit{swTVM} achieves the maximum speedup on memory-intensive operators, which mainly contain batch\_norm, relu, pooling, bias\_add operators.
These operators benefit a lot from the operator fusion, which fuses multiple small operators together to avoid redundant DMA transfers between LDM and main memory.
The batch\_norm operator in \textit{swCaffe} is implemented as a batchnorm layer (applying the mean and the variance) and a scale layer (scaling and then shifting, i.e., $ax+b$). 
\textit{swTVM} fuses mean/variance applying and scaling to the preceding convolution operator and also fuse the shifting to the next relu operator, leading to superior performance.
As for convolution and dense operators, both \textit{swTVM} and \textit{swCaffe} can leverage the optimized kernel libraries such as \textit{swDNN}, \textit{xMath}, etc., so their performance could be similar.
If these operators are configured with bias (e.g., all dense operators, convolution operators of VGG16), \textit{swTVM} regards the bias computation as memory-intensive operators while \textit{swCaffe} regards them as part of convolution/dense operators. As a result, the speedup of \textit{swTVM} on these operators are slightly better, and the speedup on memory-intensive operators may decrease slightly.
 
Besides, the convolution operators from MobileNet optimized by \textit{swTVM} show 2.74$\times$ speedup on average. 
These operators are depthwise convolutions, and each is transformed into im2col and tall-skinny matrix multiplication with parameters $M,N,K$, where $M$ is 1, $N$ represents the feature map size, $K$ is the kernel size.
In this scenario, \textit{swTVM} invokes the optimal swGEMM library rather than xMath and achieves superior performance to \textit{swCaffe}.
Although the memory-intensive operators from DCGAN has 6.82$\times$ speedup, the overall speedup of DCGAN is still negligible, as shown in Figure~\ref{fig:e2e}. It is because the memory-intensive operators is not the performance bottleneck, which contribute to less than 2\% of the end-to-end inference time.
Similarly, the memory-intensive operators from VGG16 contribute to less than 5\%.

\subsection{Roofline Analysis}
\label{subsec:roofline}
\begin{figure}
    \centering
    \includegraphics[width=0.99\linewidth]{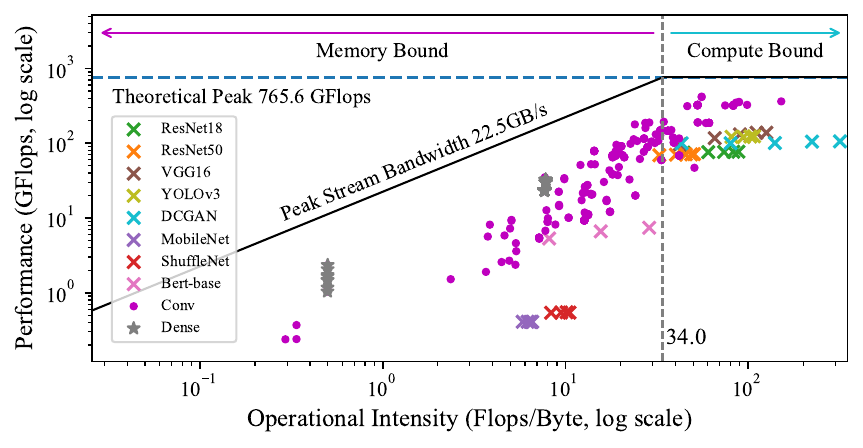}
    \caption{Roofline analysis. All benchmarks under the batch sizes of 1, 2, 4, and 8 are included.}
    \label{fig:roofline}
\end{figure}

We further perform the roofline analysis to study the effectiveness of the codes generated by \textit{swTVM}. 
Figure~\ref{fig:roofline} presents the experiment results of the overall model inference across all benchmarks, as well as the results of the convolution and dense operators.
Only the lightweight models, MobileNet and ShuffleNet, lie on the left of the ridge point. They have low operational intensity since they are designed for low-power edge devices. 
Most benchmarks lie on the right of the ridge point due to the high operational intensity and achieve better performance, because \textit{swTVM} generates efficient codes for the convolution, dense, and memory-intensive operators.
Specifically, the convolution operators optimized by \textit{swTVM} reach 419.83 GFlops, more than half of the peak performance of a CG.

\subsection{Compilation Overhead}
\label{subsec:compilation-oeverhead}
\begin{figure}
    \centering
    \includegraphics[width=\linewidth]{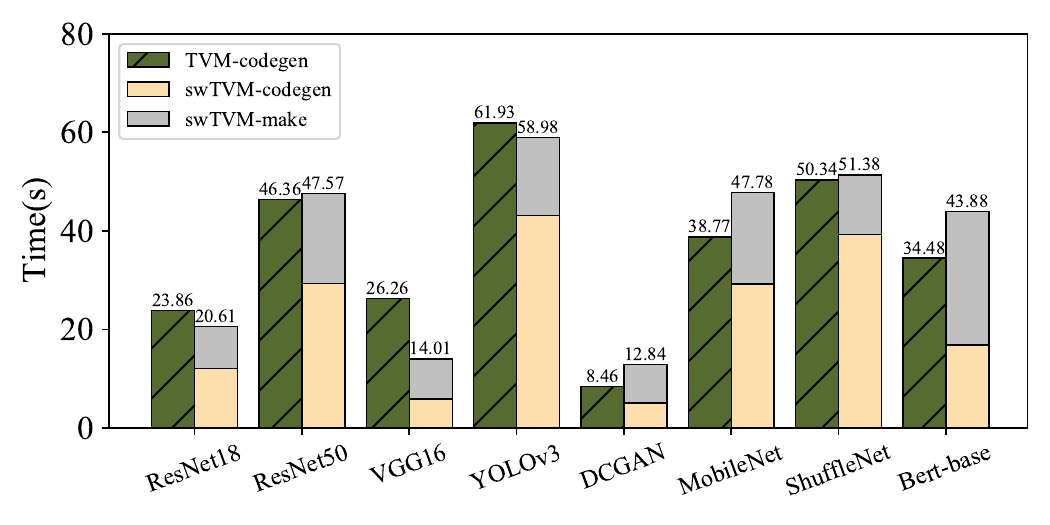}
    \caption{Compilation overhead of \textit{swTVM} on Sunway processor, comparing to that of TVM on x86 CPU.}
    \label{fig:compilation-oeverhead}
\end{figure}

The compilation overhead of \textit{swTVM} can be attributed to two parts. The first part (\texttt{codegen}) is the AOT generation of optimized C/C++ codes and corresponding makefile, whereas the second part (\texttt{make}) is the compilation through the native C/C++ compiler of Sunway. 
Figure~\ref{fig:compilation-oeverhead} presents the breakdown of the compilation overhead of \textit{swTVM} and the compilation overhead of TVM on x86 CPU.
It is obvious that their total compilation overhead are comparable.
The \texttt{codegen} time of \textit{swTVM} is much lower than TVM, whereas the \texttt{make} time is determined by the native compilers on Sunway. 

%% file: relatedwork.tex
\section{Related Work}
\label{sec:relatedwork}

\subsection{Deep Learning Compiler}
\label{subsec:compiler}
Currently, the deep learning community develops rapidly. There are always emerging deep learning models and hardware devices. However, the engineering efforts of porting various models to numerous hardware devices increase dramatically. Under this background, the end-to-end deep learning compilers are proposed.
XLA~\cite{abadi2016tensorflow} from Google focuses on the high-level computation graph, and it can fuse those subgraphs together to generate efficient code. DLVM~\cite{wei2017dlvm} is similar to TensorFlow XLA, which focuses on the high-level, but it promotes using linear algebra instead of the computation graph to express the higher-level of the models. As they pay less attention to the hardware level, significant engineering effort is needed for each hardware and operation combination. 
TVM~\cite{chen2018tvm} proposes the end-to-end compiler for neural networks and now supports various hardware. Recent works such as Glow~\cite{rotem2018glow}, Tensor Comprehensions~\cite{vasilache2018tensor} and nGraph~\cite{cyphers2018intel} can all be classified into this category. Glow lays emphasis on its two-phase strongly-typed intermediate representation and nGraph pays more attention to how to simplify the connection between deep learning frameworks and hardware. Tensor Comprehensions provides a language similar to math to describe the neural network and supports optimizing the computational kernel according to the parameter of neural networks in JIT mechanism. 
There are also a few emerging tensor compilers optimizing the bottleneck operators~\cite{akg, bladedisc, jia2019taso, wang2021pet}. 
However, they all lack the support of Sunway many-core processors.


\subsection{Performance Optimization on Sunway}
\label{subsec:optimization}
As a supercomputer consisting of massive Sunway many-cores processors, Sunway TaihuLight achieved the peak performance of 125PFlops and ranked the first place in Top500 from 2016 to 2018. There are a lot of optimization works targeting the architecture features on Sunway, which are valuable for our work to generate high performance code.

For applications, molecular dynamics~\cite{zhang2016extreme}, earthquake simulation~\cite{fu201718}, and atmospheric dynamics~\cite{7877004} won the Gordon Bell Prize of ACM. 
For algorithms, there are plenty of algorithms optimized on Sunway such as BFS~\cite{lin2017scalable}, SpMV~\cite{liu2018towards},  SpTRSV~\cite{li2018multi,swsptrsv}, and Cholesky factorization~\cite{swcholesky}. BFS is an essential algorithm in calculating the shortest route and the maximum flow problem, and the optimization on Sunway achieves 23,755 giga-traversed edges per second. Sparse computation such as SpMV is one of the important computational kernels in scientific applications. The implementation of SpMV on Sunway achieves 15.5$\times$ speedup on average over 18 representative datasets. 
There are also two related works regarding the deep learning on Sunway. \textit{swDNN}~\cite{fang2017swdnn} is a neural network library customized for Sunway with tremendous engineering efforts. \textit{swCaffe}~\cite{li2018swcaffe} proposes a deep learning framework for distributed training on Sunway.

To the best of our knowledge, there is no existing work on the end-to-end deep learning compiler that exploits the architecture advantage of Sunway processor.

%% file: conclusion.tex
\section{Conclusion}
\label{sec:conclusion}
We propose a deep learning compiler, \textit{swTVM}, for Sunway processor. 
\textit{swTVM} adopts AOT code generation to address the unique compilation environment on Sunway, and leverages several architecture features during code generation so that the computing capability of Sunway can be better utilized.
Specifically, a DMA control interface is proposed to manipulate the data access of the tensor better. A LDM management mechanism is designed to buffer data in LDM in order to reduce the memory access latency. Moreover, a DMA auto-insertion algorithm is proposed to identify the locations for inserting DMA instructions automatically with improved data re-use.
In brief, \textit{swTVM} bridges the gap of deep learning and Sunway processor with improved productivity and efficiency. 


%% file: appendix.tex
\clearpage
\appendix
\begin{algorithm}[thbp]
	\caption{LDM management algorithm.}
	\label{alg:ldm}
	\begin{algorithmic}[1] \footnotesize
		\Function{LDMManagement}{$itervars,tensorset$}
		\State /*Classify itervars to sizeiters, numiters and compiters*/
		\State $InitValue = 64$
		\State $\{sizeiters,numiters,compiters\} \gets $
		\State $\ \ \ \ \ \  CLASSIFY(itervars,tensorset)$
		\For {$ iter \in itervar $} 
			\State $\ Buffer(iter) \gets 1$
		\EndFor
		\State $Sort(compiters)$\label{alg:ldm:sort:begin}
		\State $Sort(sizeiters)$\label{alg:ldm:sort:end}
		\State $Iters = \{ sizeiters , compiters \}$
		\State /* initial buffer size */\label{alg:ldm:init:begin}
		\While {$Iters \not= \{\}$}
		\State $sizeiters \gets \{\}; compiters \gets \{\}$
		\For {$i \gets 0,LEN(Iters)$}
		\State $iter \gets Iters(i)$
		\If {$range(iter)\  \textless\  InitValue$} \label{alg:ldm:64:begin}
		\State $Buffer(iter) \gets range(iter)$
		\State $UPDATE(itervars,tensorSet,iter,UP)$
		\Else
		$\ Buffer(iter) \gets InitValue$
		\EndIf\label{alg:ldm:64:end}
		\While { $dma\_use\  \textgreater\  dma\_size$ }\label{alg:ldm:check:begin}
		\If {$Buffer(iter) == range(iter)$}\label{alg:ldm:dim:begin1}
		\State $UPDATE(itervars,tensorSet,iter,DOWN)$
		\EndIf\label{alg:ldm:dim:end1}
		\State $Buffer(iter) \gets Buffer(iter) / 2$
		\If {$Buffer(iter) == 0 $}
		\State $Buffer(iter) \gets 1; i \gets i - 2$
		\State $Break$
		\EndIf
		\EndWhile\label{alg:ldm:check:end}
		\EndFor
		\State ${sizeiters,numiters,compiters} \gets$\label{alg:ldm:dim:begin3}
		\State $\ \ \ \ \ \  CLASSIFY(itervars,tensorset)$
		\State $Sort(compiters); Sort(sizeiters)$
		\State $Iters = \{sizeiters,compiters\}$\label{alg:ldm:dim:end3}
		\EndWhile\label{alg:ldm:init:end}
		\State /* expand buffer size */
		\While {$True$}\label{alg:ldm:extent:begin}
		\State $iter = select(Iters)$
		\State $Buffer(iter) \gets Buffer(iter) * 2 $
		\If {$dma\_use\ \textgreater\  dma\_size$}
		\State $Buffer(iter) \gets Buffer(iter)/2$
		\State $Break$
		\EndIf
		\If {$Buffer(iter) == range(iter)$}\label{alg:ldm:dim:begin2}
		\State $UPDATE(itervars,tensorSet,iter,UP)$
		\State $\{sizeiters,numiters,compiters\} \gets $
		\State $\ \ \ \ \ \  CLASSIFY(iteriters,tensorset)$

		\State $Sort(compiters); Sort(sizeiters)$
		\State $Iters = \{sizeiters,compiters\}$
		\EndIf\label{alg:ldm:dim:end2}
		\EndWhile\label{alg:ldm:extent:end}
		\EndFunction
	\end{algorithmic}
\end{algorithm}

\begin{algorithm}[thbp]
	\caption{Loop reordering algorithm.}
	\label{alg:order}
	\begin{algorithmic}[1] \footnotesize
		\State /*select the iter which requires the least number of DMA transfers*/
		\Function{SELECT} {$Iters$}\label{alg:order:select:begin}
		\State $cur\_iter \gets NULL; cur\_dmatimes \gets INTMAX$
		\For {$iterid \gets 0,LEN(Iters)$}
		\State $iter \gets Iters(iterid); dmatimes \gets count(iter)$
		\If {$cur\_dmatimes \textless dmatimes$}
		\State $cur\_iter \gets iter;cur\_dmatimes \gets dmatimes$
		\EndIf
		\EndFor
		\State \Return $cur\_iter$
		\EndFunction\label{alg:order:select:end}
		\Function{ReorderLoop}{$bufferiters,iters$}
		\State $iterorder \gets [\ ]$
		\State /* Classify itervars to bufferiters and iters */
		\State $iterorder.add(bufferiters)$
		\While{true} \label{alg:order:loop:begin}
		\State $iter \gets SELECT(iters)$
		\If{$iter \not= NULL $}
		\State $iterorder.add(iter);iters.rm(iter)$
		\Else
		$\ break$
		\EndIf
		\EndWhile	\label{alg:order:loop:end}
		\State \Return $iterorder$
		\EndFunction
	\end{algorithmic}
\end{algorithm}